\documentclass{article}
\usepackage{amssymb}
\usepackage{amsmath}
\usepackage{times}
\usepackage{graphicx} 
\usepackage{subfigure} 

\usepackage{natbib}

\usepackage{algorithm}
\usepackage{algorithmic}

\usepackage{hyperref}

\usepackage[accepted]{icml2015}

\icmltitlerunning{Iterative Orthogonal Feature Projection for Diagnosing Bias in Black-Box Predictive Systems}

\begin{document} 

\twocolumn[
\icmltitle{Iterative Orthogonal Feature Projection for Diagnosing Bias in Black-Box Models}

\icmlauthor{Julius Adebayo}{juliusad@mit.edu}
\icmlauthor{Lalana Kagal}{lkagal@csail.mit.edu}
\icmladdress{CSAIL, MIT, 32 Vassar Street Cambridge, MA 02139 USA.}

\icmlkeywords{boring formatting information, machine learning, ICML}

\vskip 0.3in
]

\begin{abstract} 
Predictive models are increasingly deployed for
the purpose of determining access to services
such as credit, insurance, and employment. Despite
potential gains in productivity and efficiency,
several potential problems have yet to
be addressed, particularly the potential for unintentional
discrimination. We present an iterative
procedure, based on orthogonal projection
of input attributes, for enabling interpretability
of black-box predictive models. Through our iterative
procedure, one can quantify the relative
dependence of a black-box model on its input
attributes.The relative significance of the inputs
to a predictive model can then be used to assess
the fairness (or discriminatory extent) of such a
model.

\end{abstract} 

\section{Introduction}
\label{introduction}
Access to large-scale data has led to an increase in the use of predictive modeling
to drive decision making, particularly in industries like banking, insurance, and
employment services \cite{bryant2008big} and \cite{mayer2013big}. 
The increased use of predictive models has led to greater
efficiency and productivity. However, improper deployment of these models can lead
to several unwanted consequences. One key concern is unintentional discrimination  \cite{crawford2014big, barocas2014big}.
It is important that decisions made in determining who has access to services are, in
some sense, ‘fair.’ A predictive model can be susceptible to discrimination if it was
trained on inputs that exhibit discriminatory patterns. In such a case, the predictive
model can learn patterns of discrimination from data leading to high dependence on
protected attributes like race, gender, religion, and sexual orientation. A predictive
model that significantly weights these protected attributes would tend to exhibit disparate
outcomes for these groups of individuals. Hence, the focus of this paper is on
auditing predictive models to determine the relative significance of a model’s inputs in
determining outcomes. Given the relative significance of a model to it’s inputs, judgement can be more easily
made about the model's fairness.

The potential increased efficiency and societal gains from leveraging predictive modeling
seem limitless, and have rightly necessitated the widespread adoption of these
models. In particular, use of predictive
modeling for decision making in determining access to services is starting to
become the defacto standard in industries such as banking, insurance, housing, and
employment. As the need for more accurate forecasts or predictions has heightened, there has
been an increase in the use of complicated, often uninterpretable predictive models in
making forecasts from data. Increasingly, these predictive models tend to have millions
of parameters and are typically considered black-boxes by practitioners. This
is because the models often generate highly accurate results, but an in-depth understanding of the underlying reasons behind these accurate results is generally lacking.
Hence, practitioners resort to feeding in input to these black-box models, then generate
results without truly understanding why their models are performing well. In
fields such as computer vision or speech recognition where the task is often to identify
or recognize a signal structure, a true understanding of the internal workings of the
underlying model generating the predictions can be excused. However, in industries
such as banking, insurance, and employment where access to these services is essential
for livelihood, it is of paramount importance that the practitioner applying a
predictive model in this setting truly understands the internal workings of her model.

\section{Related Work}
\label{relatedwork}
As automated decision-making systems began to gain widespread use in rendering decisions, researchers have begun to look at the issue of fairness and discrimination in data mining. Increasingly, the emerging subfield around the topic is known as discrimination aware data mining, or fairness aware data mining \cite{pedreshi2008discrimination}. The literature on fairness is broad including works from social choice theory, game theory, economics, and law \cite{romei2014multidisciplinary}. In the computer science literature, work on identifying and studying bias in predictive models has only begun to emerge in the past few years. 

More recently, studies have started to emerge that seek to identify and correct potential bias in predictive modeling. In general, these works can be broadly classified into 3 broad categories: data transformation, algorithm manipulation, and outcome manipulation methodologies \cite{zemel2013learning, adler2016auditing}. For data transformation techniques, the input data to a data mining system is perturbed as a means of quantifying bias in the data. In \cite{calders2010three} Kamiran \& Calders present a method to transform data labels in order to remove discrimination. With their proposed method, a Naive-Bayes classifier is trained on positive labels, then a set of highly ranked negatively labeled items from the protected set are changed to achieve statistical parity of outcomes. The modified data generated is then used to learn a fairer classifier. In \cite{friedler2014certifying} Friedler et. al. also present a data ‘repair’ methodology for transforming biased data into one that predictive models can hopefully learn fair models on. In general, data transformation methodologies are typically seeking to learn fair representations of a dataset upon which less biased predictive models can be developed. 

As another class of methodologies, algorithm manipulation methods seek to augment the underlying algorithm in order to reduce discrimination. Algorithm augmentation is usually done via a penalty that adds a cost of discrimination to a model’s cost function. These algorithms typically add regularizers that quantify the degree of bias. A seminal work in this area is the study by Kamishima et. al. in \cite{kamishima2011fairness} where they quantify prejudice by adding a mutual information based regularizer to the cost function of a logistic regression model. Since the Kamishima et. al. work, more approaches that seek to change underlying cost functions with regularizers for statistical parity have emerged for other kinds of algorithms like decision trees and support vector machines. Techniques presented in this area typically only work for one particular method like logistic regression or Naive Bayes, so the overall impact can be limited. Algorithm manipulation methods also assume that underlying predictive models are known, completely specified with well-behaved cost functions. Usually, this is not the case, as a variety of models are typically combined in a number of ways where it becomes difficult to untangle the cost function for the combined model. 

In the third approach, other studies have presented work that manipulates the outcomes of predictive models towards achieving statistical parity across groups. In these cases, algorithms presented typically change the labels of data mining algorithms seeking to balance the outcomes across multiple groups. In \cite{pedreshi2008discrimination}, Pedreschi et. al. alter the confidence of classification rules inferred. 

\section{Feature Ranking Methodology}
\subsection{Problem Overview}
For this paper, the main goal of our work is to present a methodology for determining a black-box algorithm's dependence on its inputs. More specifically, given the input and output to a black-box model, we seek to produce an input ranking that corresponds to the black-box's predictive dependence on each input. We take predictive dependence as the change in performance of the black-box algorithm (defined as Mean Squared Error (regression) or Classification accuracy (Classification)).

\subsection{Orthogonal Feature Projection}
Traditionally, to make causal claims about the dependence between an input variable and a target, an experiment is needed to remove the effect of confounding variables. For example, let's assume Harvard University were running a classifier to determine its admissions decisions. If Harvard University is then accused of discriminating on the basis of gender in its admissions decisions, how would one show definitive proof of this accusation? Hypothetically, we would find two groups of applicants that are similar in all characteristics except gender, send in those applications to the university's classifier and then look at the difference in outcomes for these two groups. If the difference in outcomes between the two groups is significant, then one might conclude that the university is discriminating on the basis of gender in its admissions decisions. 

The intuition underlying the experimental process motivates our use of the orthogonal transformation for auditing a black-box model. In the above example, the experimental process is able to show the dependence of the university's classifier on race. Furthering this approach, we propose using orthogonal transformation as a tool of creating multiple copies of input data that can then be used to query a black-box model in order to determine the model's dependence on its input. Now we proceed to our overview of the iterative orthogonal transformation process. 

\subsubsection{Orthogonal Feature Projection}
Orthogonal projection is a particular type of a larger class of linear transformations. Intuitively, given two vectors whose inner product is zero, one can conclude that no linear transformation of one vector can produce the other. 

\begin{algorithm}
\textbf{INPUT:} An $n$ x $k$ data matrix $X_{pre}$ where $\vec{x_1}, \vec{x_2}, \ldots, \vec{x_k}$ represent attribute vectors for $n$ samples.\\ Current feature is $\vec{x_1}$\\ 
\textbf{OUTPUT:} An $n \times k-1$ transformed matrix $X_{new}$ that can be decomposed into $\vec{x^*_{2}}, \vec{x^*_{3}}, \ldots, \vec{x^*_{k}}$ where each vector $x^*_{i} \in X_{new}$ is orthogonal to current feature $\vec{x_1}$.
\begin{algorithmic}
\STATE Remove current attribute vector $\vec{x_1}$ from $X_{pre}$ returning $X_{del}$ 
\STATE Initialize an $n \times k-1$  vector $X_{new}$ 
\FOR{each feature $\vec{x_{i}}$ in $X_{del}$}
\STATE obtain $x^*_{i}$, the component of $\vec{x_{i}}$ that is orthogonal to current attribute vector $\vec{x_1}$ 
\STATE where $x^*_{i}$ = $\vec{x_i} - (\frac{\vec{x_1} \cdot \vec{x_i}}{\vec{x_1} \cdot \vec{x_1}}) \vec{x_1}$
\STATE join $x^*_{i}$ column wise to $X_{new}$
\ENDFOR
\STATE Return $X_{new}$
\end{algorithmic}
\caption{Linear Feature Transformation Algorithm}
\label{algo:relgraph}
\end{algorithm}

Our iterative process involves transforming each input feature iteratively in a given dataset to obtain several different copies of the dataset where each feature has been `removed.' By removed, we mean that all the other attributes in a given input have been made orthogonal to an attribute of interest. Now, given each new transformed dataset, the change in performance of the black-box can then be detected given each transformed data and used as a ranking for each feature. Below we detail the more general algorithm. Given that the above orthogonal transformation is a linear one, and most black-boxes would tend be non-linear, for each feature we augment the input matrix with additional transformations for of each feature that are non-linear in order to capture non-linear dependencies that might be present in the black-box learning algorithms. 

Note that for the General Ranking Framework presented above, we assume that we can query the black-box iteratively in order to obtain a change in predictive performance given each transformation of the input feature vector. This requirement is not always fulfilled. There are cases where only a single output from the black-box is available and the actual algorithm cannot be queried iteratively. In these cases, we learn our own equivalent representations of the black-box in form of classifiers or regressors. As we show in the evaluation sections, learning our own classifiers is subject to model misrepresentation problems as traditionally expected because there is no guarantee that our new set of classifiers are good representations of the black-box model. 
 
 \begin{algorithm}
\textbf{INPUT:} An $n$ x $k$ data matrix $X$ that can be decomposed into $\vec{x_1}, \vec{x_2}, \ldots, \vec{x_k}$ attribute vectors\\ Output of the black-box algorithm is $y$\\ Initial baseline predictive performance $b$ of the black-box algorithm. \\
\textbf{OUTPUT:} Vector $R$ $\in$  $\mathbb{R}^k$ of predictive dependence of the black-box on each input feature 
\begin{algorithmic}
\FOR{each attribute $\vec{x_{i}}$ in $X$} 
\STATE Combine non-linear transformations (log, polynomial, exponential etc) of each attribute $\vec{x_i}$ with vector $X$ as $X_{poly}$
\STATE obtain $X_{new}$ from the Feature Transformation Algorithm given $X_{poly}$
\STATE obtain black-box's predictive performance (MSE or classification accuracy) given $X_{new}$ as $b_{new}$
\STATE predictive dependence on $\vec{x_{i}} = \mid b - b_{new} \mid $  
\STATE store $\vec{x_{i}} = \mid b - b_{new} \mid $   in $R$
\ENDFOR
\STATE Return $R$
\end{algorithmic}
\caption{General Ranking Framework}
\label{algo:relgraph}
\end{algorithm}
 
 \section{Variable Case Studies}
 In this section, we demonstrate our ranking methodologies on real world data set in order to demonstrate how we expect our proposed approach to be used. We are leverage information from a major bank in Europe that developed an internal algorithm for calculating customer credit limit. The credit limit model is critical to the bank's revenue and also determine how the bank treats its customers. Suppose regulators in the Bank's region get complaints that the bank is discriminating on the basis of gender, we show our one might leverage the algorithms proposed to audit the bank's algorithm. 
 
Our dataset set in this case is demographic information for 400 thousand customers for the bank, as well as output values indicating each individual's credit limit as calculated by the bank's model. Leveraging our algorithms, we are able to rank the inputs to the bank's model in order to quantify the dependence of the bank's credit limit model on its various inputs. 
 
\begin{figure*}
\centering \includegraphics[scale=0.42]{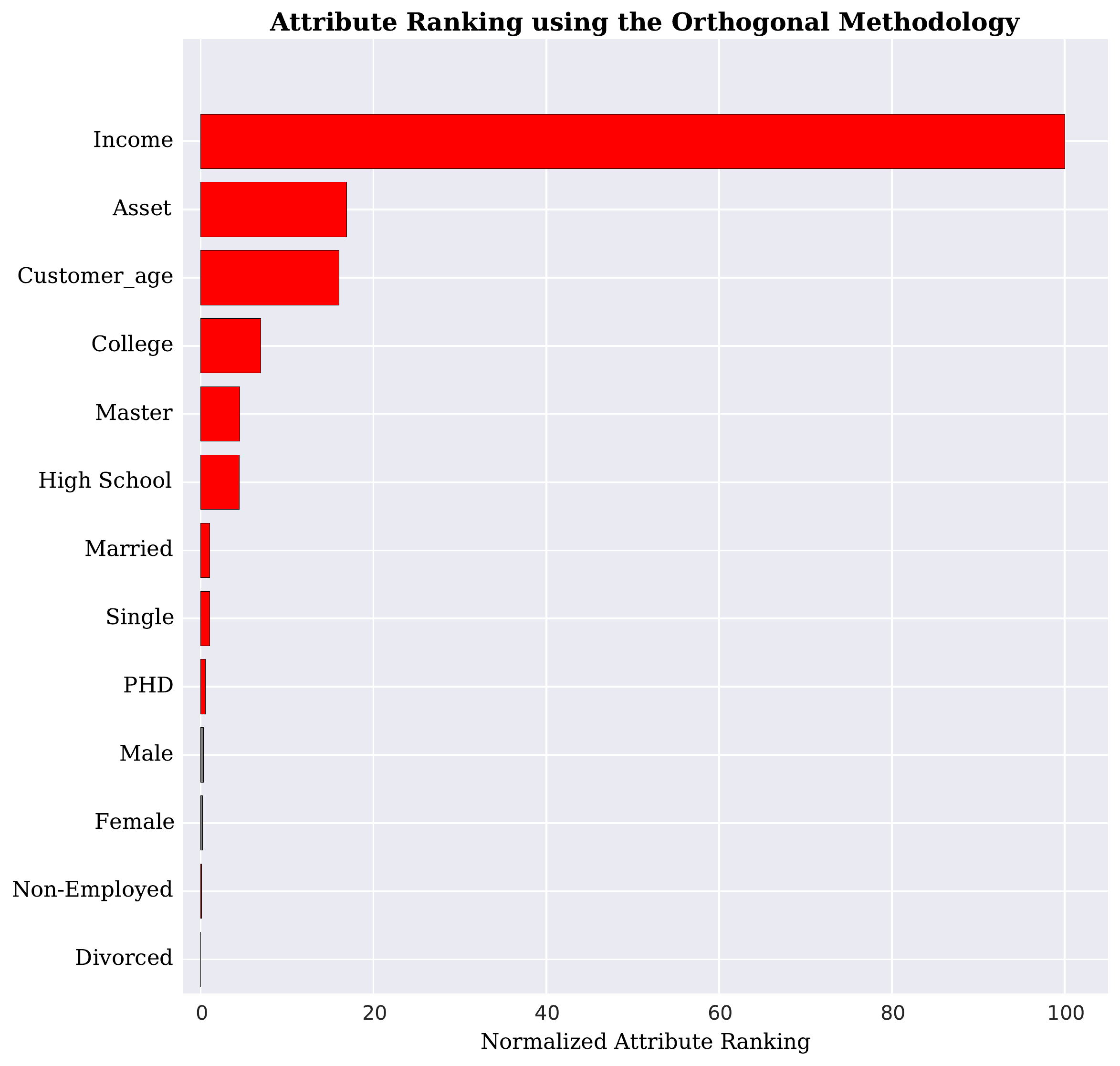}
\label{credit_limit}
\caption{Figure shows the ranking derived from our iterative orthogonal projection algorithm. The rankings have been normalized so that the most significant variable is scaled to 100 and the others relative to the most significant one. }
\end{figure*}

As we see in each of the rankings produced, the predictive dependence of the bank's credit limit algorithm on gender is consistently low across the board for all the three different ranking algorithms indicating that the bank's algorithm is not overly dependent on gender in making credit limit determinations. With simple bar plots like those shown in figure 4, we hope analysts can easily interpret the results from FairML in order to make determinations about how to investigate a particular system and what to focus on. We also ran these ranking analysis on the bank's algorithm for calculating probability of default. Due to space concerns, we have included this plot in the appendix of this document.

 \section{Conclusion and Future Work}
In this paper, we have presented an overview, and a case
study demonstrating an approach for ranking the inputs to
black-box algorithms. This work is being developed as part
of a larger project, FairML, which is a toolbox to automatically
enable interpretability of black-box models. In
this paper, we have presented our orthogonal transformation
methodology for determining a black-box algorithm’s
dependence on its inputs. Ultimately, we hope to contribute
to the large body of work regarding how to develop methods
to audit black-box machine learning systems.
 
\section{Citations and References}
\nocite{langley00}

\bibliography{example_paper}
\bibliographystyle{icml2015}

\end{document}